\documentclass[journal]{IEEEtran}

\usepackage[T1]{fontenc}
\usepackage{amsmath,amssymb}
\usepackage{bm}
\usepackage{graphicx}
\usepackage{booktabs}
\usepackage{array}
\usepackage{multirow}
\usepackage{siunitx}
\usepackage{url}
\usepackage[hidelinks]{hyperref}

\newcommand{\CVw}{\ensuremath{\mathrm{CV}_{w}}}
\newcommand{\ICC}{\ensuremath{\mathrm{ICC}(1,1)}}
\newcommand{\sigw}{\ensuremath{\sigma_{w}}}
\newcommand{\sigb}{\ensuremath{\sigma_{b}}}

\begin{document}

\title{Repeatability Characterisation and Error Budget of a\\
Consumer Structured-Light Scanner for 3-D Wound Geometry:\\
A Rigid-Phantom Study}

\author{Pushkal~Kumar,
        Aadit~Aggarwal,
        and~Karlen~Aleksanyan
%
%
%
%
\thanks{Manuscript submitted \today.
\emph{(Corresponding author: Pushkal Kumar.)}}
\thanks{This work involved no human subjects and no animal subjects. All
measurements were made on commercially manufactured synthetic wound-care
training models, so institutional review board approval and informed consent
were not applicable.}
\thanks{P. Kumar, A. Aggarwal, and K. Aleksanyan are independent researchers in
the USA (e-mail: pushkalku@gmail.com; aadit.aggarwal@gmail.com;
karlen.aleksanyan2007@gmail.com).}
\thanks{The authors declare no conflict of interest. This work received no
funding. The authors have no financial or personal relationship with VATA, Inc.,
Shining3D, or Revopoint, whose products are evaluated here.}
\thanks{Analysis code and the derived descriptor tables that reproduce every
number in this paper are available from the authors.}}

\markboth{IEEE Sensors Journal}%
{Kumar \MakeLowercase{\textit{et al.}}: Repeatability and Error Budget of a Structured-Light Scanner}

\maketitle

\begin{abstract}
We characterise the measurement error of a free-hand consumer structured-light
scanner used to derive three-dimensional geometric wound descriptors. Rigid
wound-care phantoms cannot change, so every difference between repeat scans of
one site is measurement error; all repeats come from a single scanner unit, nine
sites and 23 scans, so this characterises one instrument. The 95 percent
repeatability limit for reconstructed surface area is a factor of 4.8, with a
confidence interval from 3.0 to 7.0, so rescanning an unchanged site can shift
the reading from a 79 percent decrease to a 377 percent increase. Forty-four of
45 descriptors fall below an intraclass correlation of 0.50, and none of 248
descriptor and pipeline combinations reaches 0.75. An error budget formed by
holding the analysis region fixed, leaving sensor and reconstruction untouched,
bounds the share of variance attributable to how much surface the operator
captured at 75 percent for surface area, 91 percent for hull area and 95 percent
for bounding-box diagonal; only hull volume is majority instrumental, at 48
percent, so a better sensor would buy little. No wound is delineated anywhere in
the chain, so the comparison against the four-week area reduction used clinically
to predict healing, a ratio near 2.1, is a lower bound on the noise an
unsegmented pipeline must overcome, not a measurement of wound-area
reproducibility. Standardising the analysis region cuts the limit to 2.13,
meeting that ratio rather than clearing it. Statistical outlier removal imposes a
measured systematic area deficit near 11 percent.
\end{abstract}

\begin{IEEEkeywords}
Biomedical measurement, error analysis, measurement uncertainty,
structured light, surface reconstruction.
\end{IEEEkeywords}

\IEEEpeerreviewmaketitle

\section{Introduction}
\IEEEPARstart{C}{hronic} wounds are a large and growing burden. Recent Medicare
analyses put the affected population at 10.5 million beneficiaries, with annual
expenditure in the tens of billions of dollars \cite{carter2023chronic,
nussbaum2018economic, sen2025burden}. Because treatment decisions hinge on
whether a wound is closing, measurement is not incidental to wound care; it is
the mechanism by which care is steered.

The measurement in routine use is poor. Ruler-based length $\times$ width
overestimates area by roughly \SI{40}{\percent} relative to digital planimetry
\cite{rogers2010digital}, and its inter-rater intraclass correlation has been
reported as low as 0.34 against 0.77 for a smartphone planimetry application
\cite{seat2017reliability}. Margin identification, rather than the tracing
itself, dominates the error \cite{flanagan2003wound}, and systematic review has
found most instrument evaluations methodologically weak
\cite{omeara2012systematic}.

This has motivated a large body of work on three-dimensional capture. Structured
light, stereophotogrammetry, and more recently consumer smartphone depth sensors
have all been applied to wounds \cite{plassmann1998mavis, krouskop2002noncontact,
jorgensen2018validation, song2023lidar}, and the derived quantities have expanded
well beyond area into volume, depth distribution, surface curvature, and
roughness indices treated as digital biomarkers of healing.

The evidence for that expansion is thinner than its adoption suggests. Area
measurement by 3-D systems is reliably good, with intraclass correlations above
0.95 across several studies. Depth and volume are a different matter: commercial
3-D cameras have been reported to underestimate volume against water displacement
by 23 to \SI{58}{\percent} \cite{williams2017three}, clinical depth agreement can
fall to $R=0.49$ \cite{lasschuit2021digital}, and successive systematic reviews
conclude that 3-D wound technologies remain under-evidenced rather than
established \cite{jorgensen2016methods, tan2023portable, mehl20253d}.

A descriptor is useful for monitoring only if a clinically meaningful change in
it is larger than the noise with which it is measured. That comparison is rarely
made, in part because separating measurement error from biological change in
patients requires the biology to hold still, which it does not. Our design
sidesteps the problem: we scan rigid synthetic wound models. Their geometry
cannot change between captures, so the entire observed spread across repeat scans
is attributable to the acquisition-to-descriptor chain.

This paper makes four contributions.

\begin{enumerate}
\item A repeatability characterisation of 45 geometric wound descriptors computed
      from consumer structured-light scans of rigid phantoms, with a variance
      decomposition separating between-site morphological spread from
      within-site measurement error.
\item Per-descriptor noise floors -- repeatability coefficients and minimum
      detectable changes -- benchmarked directly against the area-reduction
      thresholds used in wound care.
\item A curvature estimator that is resolution-independent over a measured window, motivated by the observation
      that repeat scans of one site differ by up to a factor of 15 in vertex
      density, which makes per-vertex discrete curvature incomparable between
      captures.
\item An error budget apportioning the measurement variance between capture
      extent and the instrument. Capture extent accounts for \SI{75}{\percent} of
      the variance in surface area and \SI{95}{\percent} in bounding-box
      diagonal, so a better sensor would buy little; standardising the analysis
      region recovers most of it in software, cutting the surface-area
      repeatability limit from 4.6 to 2.1 while raising intraclass correlation
      from 0.02 to 0.20.
\end{enumerate}

The headline result is cautionary, and we think a useful one: it is specific, it
is quantitative, and it supplies the noise floors that a positive claim in this
area has to clear. A cautionary headline is not the whole of it, though. The same analysis identifies
which error source dominates and shows that one inexpensive change to the
processing pipeline removes most of it.

\section{Materials}

\subsection{Sensor Taxonomy}
Terminology in this literature is inconsistent. Active optical 3-D sensing
divides into time-of-flight ranging and triangulation, in which depth follows
from the geometry of a projected pattern observed off-axis. Both scanners used here are triangulation devices employing
infrared structured light. Neither is a LiDAR, and the Apple TrueDepth camera
often described as one in wound literature is likewise an infrared
dot-projector structured-light system; the rear-facing Apple LiDAR Scanner is a
genuinely distinct direct time-of-flight device with roughly centimetre accuracy
and a reported object detection limit near \SI{5}{\centi\meter}
\cite{luetzenburg2021evaluation, vogt2021comparison}. We use ``structured light''
throughout.

\subsection{Scanners}
Two commercial handheld scanners were used. The Shining3D Einstar is an infrared
VCSEL structured-light scanner with a vendor-stated point distance of 0.1 to
\SI{3}{\milli\meter} and a working distance of 160 to \SI{1400}{\milli\meter}.
The Revopoint RANGE is a binocular infrared structured-light scanner with a
vendor-stated single-frame accuracy of \SI{0.3}{\milli\meter}, a point distance
of \SI{0.3}{\milli\meter}, and a working distance of 300 to
\SI{800}{\milli\meter}.

Two points deserve emphasis because they bound what follows. First, vendors
specify point distance, a sampling density, far more often than accuracy, and the
two are routinely conflated. Second, both devices are marketed for objects
substantially larger than a wound: the Einstar documentation suggests a minimum
object of roughly \SI{100}{\milli\meter} on a side, and the RANGE specifies a
minimum scan volume of $50\times50\times\SI{50}{\milli\meter}$. Typical wounds
approach or fall below these bounds, so this study operates near the edge of the
declared operating envelope. That is not a flaw in the experiment; it is the
condition under which such devices are actually being applied to wounds.

\subsection{Phantoms}
All targets were commercially manufactured wound-care training models from
VATA Inc.\ (Canby, OR, USA): the Seymour II wound care model (sacral and hip
pressure injuries), the Wilma Wound Foot, the Vinnie Venous Insufficiency Leg,
the Annie Arterial Insufficiency Leg, and the Pat Pressure Injury Staging Model,
together with separate moulded staging reference blocks. Between them these
models present staged pressure injuries, unstageable eschar, neuropathic and
arterial ulcers, dehisced surgical wounds, and callus and fissure features.

Using phantoms is what makes the variance decomposition identifiable. The models
are rigid and unchanging, so a difference between two scans of one site is
measurement error by construction, with no biological component to disentangle.
The models are also catalogue items, so the reference geometry is available to
any group wishing to replicate this work -- a property patient wounds do not
have. Phantoms have been used for exactly this purpose in prior 3-D wound work
\cite{filko2023phantom}. The corresponding limitation, that moulded polymer
differs from living tissue in specular and subsurface optical behaviour, is
treated in Section~\ref{sec:limitations}.

\subsection{Corpus}
The corpus comprises 66 triangle meshes: 26 from the Einstar, each of a distinct
wound site, and 40 from the RANGE, covering 26 sites. Nine of the RANGE sites
were captured two or three times, giving 23 scans that constitute the
repeatability set. Scans were acquired free-hand, which is how these devices are
used in practice and is the acquisition mode whose variance we intend to
characterise. Only the two staging reference blocks were captured on both
devices, which is too few for a cross-device agreement analysis; we therefore
make no such claim.

No usable acquisition timestamps exist for this corpus, so it supports no
longitudinal analysis, and none is attempted.

\section{Methods}

\subsection{Preprocessing}
Each mesh was cleaned by statistical outlier removal followed by Taubin
smoothing. For outlier removal, the mean distance $\bar d_i$ from each vertex to
its $k=20$ nearest neighbours was computed, and vertices with
$\bar d_i > \mu_d + 2\sigma_d$ discarded.

One implementation detail is worth stating because it is easy to get wrong and
its failure is silent. A vertex cannot be deleted without also deleting every
face incident on it. Retaining those faces re-indexes them onto surviving
vertices and stitches the mesh across the resulting holes, which in our corpus
inflated surface area by up to a factor of seven and Gaussian curvature by
several orders of magnitude, while leaving a mesh that loads and renders without
complaint. A face is therefore kept only when all three of its vertices survive.

Statistical outlier removal carries a systematic cost that we quantified only
after the fact, and it is larger than we expected. Run on analytic surfaces that
contain no noise and no outliers at all, the $\mu_d + 2\sigma_d$ criterion still
discards \SIrange{4.0}{7.3}{\percent} of vertices and \SIrange{6.1}{12.7}{\percent}
of true surface area, because the threshold is relative and therefore always
removes an upper tail. Over the full chain the reconstructed area of an object of
known geometry comes out about \SI{11}{\percent} low. Every absolute area in this
paper carries that deficit. It does not affect any conclusion here, because every
headline statistic is a ratio of two areas measured through the same chain and the
bias is common-mode, but it does mean the absolute areas should not be read as
calibrated measurements.

Smoothing used the Taubin $\lambda\mid\mu$ filter with $\lambda = 0.5$,
$\mu = -0.53$, and ten iterations \cite{taubin1995siggraph, taubin1995iccv},
which alternates a shrinking and an inflating Laplacian step so that low
frequencies pass while high-frequency noise is attenuated. Section~\ref{sec:smoothing}
reports the measured effect of this choice rather than assuming it.

\subsection{Descriptors}
Every scan in this corpus is an open surface: none of the 66 meshes is
watertight. This immediately invalidates a family of descriptors in common use.
Enclosed mesh volume is undefined on an open surface, and so is the
hull-minus-mesh difference frequently used as an ``estimated wound volume.'' We
compute convex hull volume, which is well defined, and report enclosed volume as
missing rather than substituting a number that has no geometric meaning.

\subsubsection*{What the descriptors are computed over}
One definitional point governs how every number in this paper should be read. No
wound is segmented anywhere in our pipeline. Surface area is the area of the
reconstructed mesh, which is whatever the operator framed into the capture --
regions of 65 to \SI{359}{\centi\meter\squared} in the repeat set, often an
entire limb or sacral region rather than an isolated wound bed. Clinical
planimetry, by contrast, measures a delineated wound margin, typically 1 to
\SI{20}{\centi\meter\squared}.

These are different measurands, and we do not claim otherwise. What we
characterise is the reproducibility of descriptors as computed in automated
pipelines of this kind, where the mesh is taken as given and no delineation
intervenes; clinical comparisons below are stated as a lower bound on the noise
such a pipeline must overcome. Section~\ref{sec:delin} reports what happens when
a delineation step is added.

We extract 45 numeric descriptors in two families. \emph{Extensive} descriptors
scale with how much surface was captured: surface area, hull area and volume,
bounding-box dimensions and diagonal, and vertex count. \emph{Intensive}
descriptors are ratios or per-unit quantities intended to be scale-free: shape
ratios, sphericity, flatness, bounding-box fill, and distributional statistics of
mean and Gaussian curvature (mean, standard deviation, median, four percentiles,
interquartile range, skewness, kurtosis), plus a roughness coefficient of
variation, root-mean-square curvature, and concave fraction.

Per-vertex curvature was estimated with the cotangent Laplace--Beltrami operator
for mean curvature and the angle-deficit formula for Gaussian curvature
\cite{meyer2003discrete, pinkall1993computing}. Boundary vertices are excluded:
both estimators are undefined on an open boundary, and including them injects a
large artefact that scales with how the scan happened to be cropped.

\subsection{Scale-Normalised Curvature}
\label{sec:scalecurv}
Discrete per-vertex curvature is defined on the mesh graph, so its value depends
on how finely the surface is sampled. In this corpus that dependence is not a
subtlety: repeat scans of a single site differ by a median factor of 4.6 in
vertex density, and density ranges over a factor of 43 across the corpus. That
spread sits almost wholly inside one instrument: the Revopoint scans span a
factor of 43 between them while the Einstar scans span 1.3, so the confound
lives within the device on which every repeatability result here is computed.
Two scans of the same rigid wound therefore yield curvature statistics computed
at different effective scales.

We remove that dependence by estimating curvature at a fixed spatial scale
instead of per vertex. A fixed number of points $N_s = 4000$ is sampled uniformly
by area; around each, all vertices within a fixed radius $r$ are collected and a
quadric
\begin{equation}
z = a x^{2} + b xy + c y^{2} + d x + e y + f
\end{equation}
is fitted by least squares in a local frame whose third axis is the
smallest-variance direction of the neighbourhood. Mean and Gaussian curvature
follow from the first and second fundamental forms,
\begin{equation}
K = \frac{LN - M^{2}}{EG - F^{2}},
\qquad
H = \frac{EN - 2FM + GL}{2\left(EG - F^{2}\right)},
\end{equation}
with $E = 1 + d^{2}$, $F = de$, $G = 1 + e^{2}$ and
$L,M,N$ the second-form coefficients obtained from $a,b,c$. Because both the
support radius and the sample count are held constant, the estimator no longer
inherits mesh resolution. That independence holds over a window rather than by
construction, and Section~\ref{sec:resol} measures where the window ends. We
evaluate $r \in \{2,3,5\}\,\si{\milli\meter}$.

\subsection{Variance Decomposition}
In the vocabulary of ISO 5725-1 \cite{iso5725}, what we estimate is a
precision statistic under conditions between repeatability and intermediate
precision: the same operator, procedure and instrument, on an unchanging
measurand, with the scan repeated. Trueness is not estimated, since no certified
artefact was available; Section~
\ref{sec:limitations} returns to that.

For a descriptor measured on $n$ sites with $m_i$ repeats each, a one-way
random-effects model gives the within-site variance $\sigw^2$, which is
measurement error since the phantoms are rigid, and the between-site variance
$\sigb^2$, which is morphological spread across distinct wounds. The intraclass
correlation
\begin{equation}
\ICC = \frac{\sigb^{2}}{\sigb^{2} + \sigw^{2}}
\end{equation}
is the fraction of observed variance attributable to morphology. We use the
one-way form because each site is measured by the same procedure with no crossed
rater factor, and we report exact $F$-based confidence intervals following the
reporting guidance of Koo and Li \cite{koo2016icc, shrout1979intraclass}.

\subsection{Precision and Detectable Change}
\label{sec:precision}
Intraclass correlation answers whether a descriptor separates \emph{these}
wounds, which depends on the cohort. For monitoring one wound over time the
relevant question is different and more portable: how large must a change be
before it exceeds noise? That is set by within-site precision alone. We report
the pooled within-site coefficient of variation \CVw, and the repeatability
coefficient $\mathrm{RC} = 2.77\,\sigw$, the 95\,\% limit on the difference
between two measurements of an unchanged wound \cite{blandaltman1986,
blandaltman1999}.

Most of these descriptors are ratio-scale quantities whose error is proportional
rather than additive, and several carry $\CVw$ above \SI{50}{\percent}, where a
symmetric limit expressed as a percentage is both distributionally implausible
and awkward to interpret -- a bounded quantity cannot fall by more than
\SI{100}{\percent}. The primary limits of agreement are therefore computed on
log-transformed values and back-transform, following the standard treatment for
proportional error \cite{blandaltman1999}. The result is a multiplicative
repeatability ratio
\begin{equation}
R_{95} = \exp\!\left(2.77\,\sigma_{\log}\right),
\end{equation}
read as ``two scans of an unchanged wound can differ by up to a factor of
$R_{95}$.'' Confidence intervals for both forms come from a cluster bootstrap
over sites with 2000 resamples, since repeats within a site are not independent.

Because \ICC{} and \CVw{} can diverge -- a descriptor that is nearly constant
across wounds is precise but uninformative -- we also report the signal-to-noise
ratio $\sigb/\sigw$, and treat a descriptor as potentially useful only when it is
both precise and varies more between wounds than within them.

\section{Results}

\subsection{Capture Variability}
\begin{figure}[!t]
\centering
\includegraphics[width=\columnwidth]{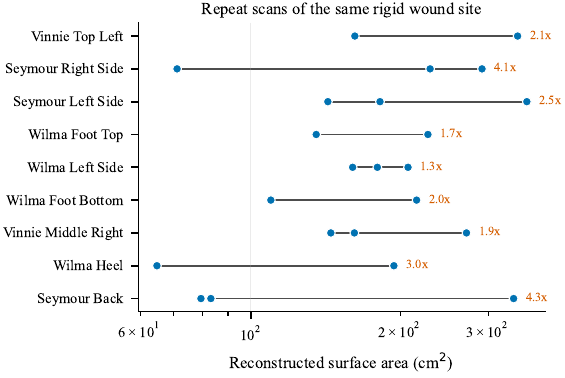}
\caption{Reconstructed surface area for every repeat scan of each site. The
phantoms are rigid, so all spread within a row is measurement error. Annotations
give the max/min ratio within each site. Note the logarithmic axis.}
\label{fig:capture}
\end{figure}

Repeat scans of one site vary enormously in what they capture
(Fig.~\ref{fig:capture}). The median within-site ratio between the largest and
smallest vertex count is 3.5, reaching 17.1 at one site; the median vertex
density ratio is 4.6; and the median ratio of reconstructed surface area is 2.12.
A twofold difference in the measured surface area of a rigid object between two
captures, before any descriptor is computed, sets the scale of the problem
(Fig.~\ref{fig:capture}).

\subsection{Precision and Detectable Change}
\begin{figure}[!t]
\centering
\includegraphics[width=\columnwidth]{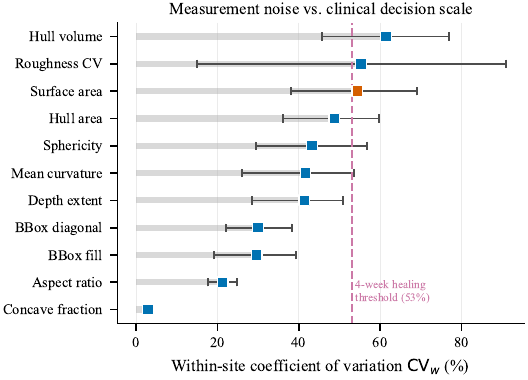}
\caption{Within-site coefficient of variation with cluster-bootstrap 95\,\%
confidence intervals, against the four-week area reduction used clinically to
predict healing. Every descriptor except the near-constant concave fraction sits
at or approaching that decision scale.}
\label{fig:precision}
\end{figure}

Fig.~\ref{fig:precision} gives within-site precision
for the descriptors of greatest clinical interest.

Surface area has $\CVw = \SI{54.4}{\percent}$
(95\,\% CI 38.1--69.0); the median difference between two scans of the same
unchanged wound is \SI{60.8}{\percent}, and the minimum detectable change is
\SI{151}{\percent}. Convex hull volume is worse, at $\CVw = \SI{61.4}{\percent}$
and $\mathrm{MDC} = \SI{170}{\percent}$. A minimum detectable change above
\SI{100}{\percent} is a signal that the symmetric parameterisation has been
pushed past its useful range, which is why we treat the log-scale limits below as
primary; the two are the same quantity expressed on different scales.

On the log scale, which is the appropriate one for these proportional errors,
the 95\,\% repeatability limit for surface area is a factor of
$R_{95} = 4.77$ (95\,\% CI 3.02--6.95), so rescanning one unchanged wound can
shift the reading by anywhere from $-79$ to $+\SI{377}{\percent}$. Convex hull
volume is far worse at $R_{95} = 10.6$ (95\,\% CI 4.12--28.19), and only the
near-constant concave fraction is tight, at $R_{95} = 1.08$
(Table~\ref{tab:logrep}).

\begin{table}[!t]
\caption{Multiplicative 95\,\% Repeatability Limits $R_{95}$ With
Cluster-Bootstrap Confidence Intervals. A Value of 1.0 Would Mean Perfect
Repeatability.}
\label{tab:logrep}
\centering
\begin{tabular}{lrr}
\toprule
\textbf{Descriptor} & $\bm{R_{95}}$ & \textbf{95\,\% CI} \\
\midrule
Concave fraction      & 1.08  & 1.05--1.11 \\
Bounding-box diagonal & 2.47  & 1.86--3.31 \\
Mean curvature        & 3.48  & 2.12--5.31 \\
Depth extent          & 3.90  & 2.39--5.72 \\
Hull area             & 4.75  & 2.79--7.92 \\
\textbf{Surface area} & \textbf{4.77}  & \textbf{3.02--6.95} \\
Roughness CV          & 5.07  & 1.55--13.76 \\
Hull volume           & 10.63 & 4.12--28.19 \\
\bottomrule
\end{tabular}
\end{table}

These figures are best read against the thresholds clinicians actually use. A
\SI{53}{\percent} reduction in wound area at four weeks is the classical
predictor of healing at twelve weeks in diabetic foot ulcers, with an analogous
threshold near \SI{44}{\percent} for venous leg ulcers. A \SI{53}{\percent}
reduction is a ratio of 0.47, that is, a factor of 2.13. The measurement process
alone spans a factor of 4.77, so the noise band is 2.2 times wider than the
change the clinician is trying to detect.

We state this comparison carefully. The descriptors are computed over the
reconstructed patch rather than a delineated wound bed, so this is not a
measurement of wound-area reproducibility; our limit comes from scans separated
by minutes against a four-week clinical threshold; and five of the nine repeat
sites were captured within one session, so the estimate blends within- and
between-session conditions.

What the comparison does establish, and all it establishes, is a lower bound on
the noise that an unsegmented pipeline of this kind must overcome before it can
resolve the effect clinicians act on. That bound exceeds the effect. The
region-standardisation result in Section~\ref{sec:remedies} shows what closes
the gap.

Only one of 45 descriptors achieves a minimum detectable change below
\SI{50}{\percent}. That descriptor is the concave fraction, and it fails for the
opposite reason: it is precise ($\CVw = \SI{2.9}{\percent}$) because it is nearly
constant, ranging only from 0.49 to 0.62 across all 66 scans. Roughly half of any
noisy surface is locally concave, so the quantity is close to a property of
surface noise rather than of the wound. Its signal-to-noise ratio is 0.57.

\subsection{Variance Decomposition}
Consistent with the precision results, 44 of 45 descriptors fall below an
intraclass correlation of 0.50 (Table~\ref{tab:icc}), the
conventional boundary of poor reliability, and none reaches 0.75.

The best descriptor is mean
curvature at $\ICC = 0.560$, with a confidence interval from 0.103 to 0.867 that
illustrates how little nine sites constrain the estimate. Extensive descriptors
are worse than intensive ones by an order of magnitude in mean \ICC{} (0.016
against 0.195), which is what one expects when capture extent varies twofold and
the descriptor scales with it.

\begin{table}[!t]
\caption{Intraclass Correlation by Descriptor Family and Processing Pipeline.
No Descriptor Reaches the Good-Reliability Threshold of 0.75 Under Any Pipeline.}
\label{tab:icc}
\centering
\footnotesize
\begin{tabular}{@{}l@{\hspace{5pt}}c@{\hspace{5pt}}c@{\hspace{5pt}}c@{\hspace{5pt}}c@{}}
\toprule
\textbf{Pipeline} & $\bm{n}$ & \textbf{Mean} & \textbf{Max} & $\bm{\geq 0.75}$ \\
\midrule
As-acquired            & 45 & 0.148 & 0.560 & 0/45 \\
\quad extensive       & 12 & 0.016 & 0.068 & 0/12 \\
\quad intensive       & 33 & 0.195 & 0.560 & 0/33 \\
ROI $r{=}15$\,mm       & 41 & 0.065 & 0.312 & 0/41 \\
ROI $r{=}25$\,mm       & 41 & 0.167 & 0.502 & 0/41 \\
ROI $r{=}40$\,mm       & 41 & 0.142 & 0.436 & 0/41 \\
Scale-norm. $r{=}2$\,mm & 27 & 0.108 & 0.450 & 0/27 \\
Scale-norm. $r{=}3$\,mm & 27 & 0.107 & 0.462 & 0/27 \\
Scale-norm. $r{=}5$\,mm & 26 & 0.095 & 0.452 & 0/26 \\
\midrule
\textbf{All comb.}      & \textbf{248} & \textbf{0.122} & \textbf{0.560} & \textbf{0/248} \\
\bottomrule
\end{tabular}
\end{table}

\subsection{Do the Obvious Remedies Help?}
\label{sec:remedies}
Two interventions follow directly from the diagnosis, and we tested both.

\subsubsection{Standardising the analysis region}
Cropping every scan to a fixed-radius patch about its centroid before computing
descriptors addresses capture extent directly. Judged by intraclass correlation
the effect looks modest, with mean \ICC{} for extensive descriptors rising from
0.016 to 0.204. That reading is misleading, because \ICC{} is bounded above by
the between-site spread of this particular cohort. Judged by precision, which is
what determines whether a change is detectable, the effect is large
(Table~\ref{tab:roiprec}).

\begin{table}[!t]
\caption{Region Standardisation Trades Precision Against Signal. $R_{95}$ Is the
Multiplicative Repeatability Limit (Lower Is Better); $\ICC$ and the Between-Site
CV Measure Whether Any Site-Specific Signal Survives. $A/\pi r^{2}$ Is the
Fraction of the Crop Disc the Retained Surface Fills.}
\label{tab:roiprec}
\centering
\footnotesize
\begin{tabular}{@{}l@{\hspace{5pt}}c@{\hspace{5pt}}c@{\hspace{5pt}}c@{\hspace{5pt}}c@{}}
\toprule
\textbf{Arm} & $\bm{R_{95}}$ & \CVw{} (\%) & \textbf{Between-site CV} & $\bm{\ICC}$ \\
\midrule
\multicolumn{5}{l}{\textit{Surface area}} \\
As-acquired            & 4.59  & 53.4 & --   & 0.000 \\
ROI $r{=}15$\,mm & 11.14 & 65.0 & 26.1\,\% & 0.000 \\
ROI $r{=}25$\,mm & 2.13  & 27.0 & 17.1\,\% & \textbf{0.202} \\
ROI $r{=}40$\,mm & 1.63  & 17.8 & 14.7\,\% & 0.000 \\
\midrule
\multicolumn{5}{l}{\textit{Crop saturation}, $A/\pi r^{2}$} \\
$r{=}15$\,mm & \multicolumn{4}{c}{0.685} \\
$r{=}25$\,mm & \multicolumn{4}{c}{0.813} \\
$r{=}40$\,mm & \multicolumn{4}{c}{\textbf{1.002}} \\
\bottomrule
\end{tabular}
\end{table}

Read by precision alone, larger radii look progressively better: the
surface-area repeatability limit falls from 4.59 as-acquired to 2.13 at
\SI{25}{\milli\meter} and 1.63 at \SI{40}{\milli\meter}. That reading is a trap,
and it is the same trap the concave fraction fell into in
Section~\ref{sec:precision}.

At $r = \SI{40}{\milli\meter}$ the retained surface fills the crop disc to within
\SI{0.2}{\percent} ($A/\pi r^{2} = 1.002$), and the mean bounding-box diagonal is
\SI{114.4}{\milli\meter} against the disc's own \SI{113.1}{\milli\meter}. The
descriptor has stopped measuring the wound and started measuring our crop.
Intraclass correlation confirms it: 0.000 for both surface area and bounding-box
diagonal, the latter with an upper confidence bound of 0.079. An $R_{95}$ of 1.63 is then
the repeatability of a geometric convention, and we do
not offer it as evidence that anything clinical has been achieved.

The informative arm is $r = \SI{25}{\milli\meter}$, where the crop is not yet
saturated (0.813), between-site variation survives at \SI{17.1}{\percent}, and
intraclass correlation rises from 0.016 as-acquired to 0.202 -- the largest
improvement in site discrimination anywhere in this study. Its repeatability
limit is 2.13, against a clinical decision ratio of 2.13. The honest optimum
lands exactly on the threshold rather than clearing it.

At $r = \SI{15}{\milli\meter}$ the crop is too small: precision is worse than
as-acquired ($R_{95} = 11.14$) and four scans lose their region entirely. Region
standardisation is thus bounded on both sides, and its apparent best case is an
artefact.

\subsubsection{Removing the resolution dependence}
The fixed-scale quadric estimator of Section~\ref{sec:scalecurv} removes the
sampling-density dependence by construction, and we validated it against analytic
ground truth: on a sphere of radius \SI{20}{\milli\meter} it recovers
$H = \SI{0.0502}{\per\milli\meter}$ against a true \num{0.05} and
$K = \SI{0.00252}{\per\milli\meter\squared}$ against a true \num{0.0025}, with
consistent sign; on an open cylindrical patch of radius \SI{15}{\milli\meter}
it recovers $H = \num{0.0336}$ against a true $1/2R = \num{0.0333}$, and
$K \approx 0$.

It does not improve reliability. Mean \ICC{} across the curvature block is 0.107
at $r = \SI{3}{\milli\meter}$, against 0.195 for the intensive descriptors
as-acquired, and the maximum is 0.462. Making curvature comparable across
resolutions is necessary for the descriptors to mean the same thing in two
scans, but it does not manufacture signal that the acquisition did not capture.

\subsubsection{Would segmenting the wound fix it?}
\label{sec:delin}
The natural objection is that these descriptors are computed over the captured
patch rather than a delineated wound, and that segmenting first would resolve it.
We tested the automated form of that fix: a reference surface from heavy
Laplacian smoothing, per-vertex depth below it taken relative to the median, and
the largest connected patch deeper than a threshold kept as the cavity. It fails,
and in the direction the rest of this paper predicts. At a \SI{2}{\milli\meter}
threshold a cavity is found in only 12 of 23 repeat-set scans, and at all repeats
of a site in only three of nine sites; where two scans of one site both yield a
cavity, the extracted area differs by a median factor of 64.5 and by as much as
247. Against a whole-patch limit of $R_{95}=4.59$, automated delineation is about
an order of magnitude worse. The reason is the one the error budget identifies:
the reference surface is built from the captured mesh, so it inherits capture
extent, and a depth measured against it inherits it too. Segmentation does not
escape the dominant error source by operating downstream of it. Manual tracing by
trained raters uses information no geometric criterion has, and remains the
right fix.

\subsubsection{Where the fixed-scale estimator stops working}
\label{sec:resol}
Section~\ref{sec:scalecurv} introduced the fixed-scale estimator to make
curvature comparable between scans of unequal density. We measured that rather
than assuming it. The eight densest meshes were preprocessed once and then
decimated by quadric edge collapse down a ten-rung ladder from \num{16} to
\SI{0.71}{\per\milli\meter\squared}, the range the corpus spans, with every
descriptor recomputed at each rung against that mesh's own dense reference.

Size and shape-ratio descriptors are effectively resolution-free: median
absolute bias runs from \SI{0.05}{\percent} to \SI{0.32}{\percent} and from
\SI{0.10}{\percent} to \SI{0.56}{\percent} across the whole ladder, so the
precision and error-budget results above are not resolution artefacts.
Per-vertex curvature is not: median absolute bias is already
\SI{15.4}{\percent} at the densest rung and reaches \SI{63.8}{\percent} at the
sparsest. The premise behind the estimator is confirmed.

Its remedy is only partly delivered, and this corrects a claim we made earlier.
At $r = \SI{3}{\milli\meter}$ the median absolute bias sits near the
estimator's own seed-to-seed floor from \num{16} down to about
\SI{4}{\per\milli\meter\squared} (\SIrange{4.2}{5.3}{\percent}), then grows to
\SI{8.3}{\percent} at \num{2.0}, \SI{17.3}{\percent} at \num{1.0} and
\SI{40.3}{\percent} at \SI{0.71}{\per\milli\meter\squared}. A larger support
extends the window, with $r = \SI{5}{\milli\meter}$ holding within
\SIrange{2.6}{8.6}{\percent} across the entire ladder, but a larger support also
truncates genuine small-scale curvature. The estimator is therefore
resolution-independent over a measured window, not by construction, and $r$ must
be chosen jointly against the sparsest density and the smallest feature a study
needs to resolve.

\subsubsection{The combined screen}
Across all 248 descriptor--pipeline combinations, none reaches
$\ICC = 0.75$, and none combines $\CVw < \SI{25}{\percent}$ with
$\sigb/\sigw > 1$. Only two combinations have a between-site signal exceeding
their within-site noise at all: mean curvature as-acquired
($\sigb/\sigw = 1.13$, $\CVw = \SI{41.6}{\percent}$) and Gaussian-curvature
kurtosis under \SI{25}{\milli\meter} region standardisation
($\sigb/\sigw = 1.00$, $\CVw = \SI{90.8}{\percent}$). Both are too imprecise to
act on.

This is the sense in which the remedies are partial. Region standardisation
substantially improves \emph{precision}, which is what longitudinal monitoring
needs, and on that axis it is close to sufficient for surface area. Neither
remedy makes any descriptor a reliable \emph{discriminator} between wound sites
in this cohort.

\subsection{Error Budget}
\label{sec:budget}
The interventions above are not only remedies; each one controls a named error
source, so re-measuring with it applied apportions the variance. On the log scale
$\sigma_{\log} = \ln R_{95} / 2.77$ and variances add, so controlling a cause and
differencing gives its share directly,
\begin{equation}
\sigma^{2}_{\text{cause}} = \sigma^{2}_{\text{uncontrolled}} - \sigma^{2}_{\text{controlled}} .
\end{equation}

Region standardisation at \SI{25}{\milli\meter} controls capture extent while
leaving the sensor, the reconstruction, and the processing untouched. The
resulting split is lopsided. Capture extent accounts for \SI{75}{\percent} of the
measurement variance in surface area, with $R_{95}$ falling from 4.59 to 2.13
once it is controlled, \SI{91}{\percent} in hull area (4.89 to 1.61) and
\SI{95}{\percent} in bounding-box diagonal (2.51 to 1.23). Everything
attributable to the instrument itself -- sensor noise, stereo reconstruction,
meshing, and our own preprocessing -- accounts for the remaining quarter or less.
Enclosed hull volume is the exception at \SI{48}{\percent} (11.81 to 5.98), and
is the one descriptor whose error is majority-instrumental.

Two consequences follow. A better sensor would buy little: eliminating
instrument error entirely would reduce the surface-area limit only from 4.59 to
about 2.1, which region standardisation already achieves in software. Enclosed
volume is the exception, with barely half its variance attributable to extent --
it is reconstructed from a surface nowhere watertight in this corpus, and that
residual is where sensor quality genuinely binds.

\subsection{Effect of Smoothing}
\label{sec:smoothing}
We measured the effect of smoothing on 14 meshes rather than assume it
Taubin smoothing preserves bulk geometry well: after
ten iterations the median change in convex hull volume is
\SI{-0.03}{\percent} (IQR $-0.07$ to $-0.00$), against \SI{+1.29}{\percent}
(IQR $+0.49$ to $+2.84$) for plain Laplacian smoothing at the same iteration
count.

The effect on texture is another matter. Ten Taubin iterations attenuate
root-mean-square curvature by \SI{49.8}{\percent}, removing half the texture
signal. Laplacian smoothing degrades non-monotonically, with root-mean-square
curvature rising \SI{435}{\percent} at ten iterations and by two orders of
magnitude by thirty, as the filter drives triangles toward degeneracy and the
angle-deficit estimator divides by vanishing vertex areas.

This matters for any roughness descriptor: smoothing suppresses exactly the
high-frequency content roughness measures, so a roughness index is a joint
function of the wound and the smoothing parameters, and such values are not
comparable across studies that smooth differently. Our parameters are stated for that reason, and roughness claims in this
literature are probably best read with the preprocessing in view.

\section{Discussion}

\subsection{What Limits These Measurements}
The binding constraint is not sensor resolution: the scanner specifies a point
distance far finer than the effects we are trying to resolve. It is that a
free-hand operator does not capture the same surface twice, with coverage varying
twofold in area and up to seventeenfold in vertex count between repeats of one
site. Size-dependent descriptors inherit that variance directly, intensive ones
indirectly through resolution-dependent estimators and boundary effects.

This is encouraging: acquisition variance is an engineering problem with known
remedies -- fixed standoff, a registration fiducial, a defined wound margin,
resampling to fixed density -- whereas a sensor noise floor would not be. Our post hoc remedies each move the metric without being sufficient, and
automated delineation makes matters worse (Section~
\ref{sec:delin}). A protocol
that controls framing at acquisition time, rather than compensating afterwards,
is the obvious next step.

\subsection{Relation to Prior Work}
Our findings agree with the more careful strand of this literature: reviews
finding 3-D wound technology under-evidenced, only moderate accuracy for depth
and volume against high reproducibility for area, volume underestimated by 23 to
\SI{58}{\percent} against water displacement, and clinical depth correlations near
$R=0.5$ \cite{jorgensen2016methods, tan2023portable, mehl20253d,
williams2017three, lasschuit2021digital}.

The tension with studies reporting high 3-D area reliability is largely
explained by acquisition conditions: those studies use fixed or tripod-mounted
capture, a defined wound margin, and trained operators, whereas the appeal of
consumer scanners is that they dispense with that apparatus. Our
numbers describe the free-hand consumer regime, and the convenience is not free.

\section{Limitations}
\label{sec:limitations}
Nine repeat sites is a small basis for a variance decomposition, and the
intraclass correlations are the weakest part of this study. Their intervals are
so wide as to be nearly uninformative: 43 of 45 lower bounds sit at zero, and 12
of 45 upper bounds reach 0.75 or above. So the claim that no descriptor reaches
good reliability is a claim about point estimates, and the data
cannot exclude that several descriptors would clear that bar in a larger study.
We report the correlations because they are the conventional currency of
reliability work, but the precision statistics carry the argument: they depend
only on within-site spread, are estimated from 14 within-site pairs, and are
correspondingly better constrained.

The repeatability estimate mixes two conditions. Five of the nine repeat sites
(the Seymour and Vinnie sites) were captured within one session, while the four
Wilma sites span two sessions. Under ISO 5725-1 terminology \cite{iso5725} the former approximate
repeatability conditions and the latter intermediate precision, and our pooled
figure is a blend. We had expected between-session variability to dominate; it does not. The
surface-area repeatability limit is 5.57 for the five same-session sites against
3.44 for the four cross-session sites, with the same ordering for hull area, hull
volume, and bounding-box diagonal. Session is fully confounded with phantom
identity here -- same-session sites are all Seymour and Vinnie, cross-session all
Wilma -- so the contrast reflects which objects are harder to capture rather than
an isolated session effect.

The corpus excludes a third set of 25 smartphone scans, which are surfel point
clouds without mesh topology and admit none of these descriptors, so we say
nothing about smartphone capture, the modality of greatest practical interest.

Sphericity is retained despite assuming a closed surface: on these open meshes it
reaches 8.9 against a theoretical bound of 1. We keep it because it appears in
comparable studies and its instability is informative, but it is not sphericity in
the usual sense.

The design cannot fully separate sensor noise from operator framing; what it
characterises is the end-to-end chain from free-hand acquisition to descriptor,
which we take to be the clinically relevant quantity.

Moulded polymer phantoms differ optically from living tissue in specularity,
subsurface scattering and moisture, and real wounds add exudate and dressing
residue. Whether phantom repeatability is optimistic or pessimistic relative to
patients is not established here, though the absence of patient movement suggests
optimistic.

The scope of the instrument claim needs stating plainly. The corpus was captured
on two scanners, but every repeat pair is from the Revopoint RANGE; the Einstar
contributes 26 single scans and no repeats, so it enters the corpus description
and nothing else. Every repeatability, correlation, precision, region and
error-budget result in this paper therefore characterises one instrument. Only
the two staging blocks were captured on both devices, which is too few for an
agreement analysis, so we report no reproducibility across instruments and the
figures here should not be assumed to transfer to the Einstar or to any other
scanner.

We also report precision without trueness. Nothing in this corpus is a
dimensionally certified artefact, and no reference measurement -- coordinate
measuring machine, water displacement, or calibrated gauge -- was available, so
we can say how consistently these descriptors are measured but not how close they
are to the truth. A scanner that is repeatably wrong would pass every test in
this paper. Trueness would require scanning an artefact of known geometry, which
is the cheapest thing a follow-up study could add.

Finally, the corpus carries no reliable acquisition timestamps, so nothing here
speaks to longitudinal measurement, and we make no claim about healing
prediction.

\section{Conclusion}
We characterised the measurement repeatability of 45 three-dimensional geometric
wound descriptors computed from free-hand consumer structured-light scans of
rigid phantoms, a design in which all variability between repeat scans is
measurement error. The 95\,\% repeatability limit for reconstructed surface area
is a factor of 4.8, so two scans of an unchanged wound can differ by $-79$ to
$+\SI{377}{\percent}$; 44 of 45 descriptors fall below an intraclass correlation
of 0.50, and across 248 descriptor--pipeline combinations none reaches 0.75.

Capture extent, not sensor resolution, bounds these measurements, and that is the
practical content of the study, because capture extent is addressable in
software. Cropping the analysis region to a \SI{25}{\milli\meter} radius cuts the
surface-area limit to 2.13, which meets rather than clears the ratio the clinical
threshold represents, and is the only manipulation we tested that also improves
site discrimination; larger radii improve precision only by making the descriptor
a function of the crop. A fixed-scale curvature estimator, validated here
against analytic ground truth, is resolution-independent across the density
range of the repeat set but not beyond it, and adds no reliability of its own
on this cohort.

Descriptors of this kind should therefore be reported with the noise floors that
bound them, and healing-prediction claims built on them checked against those
floors first.

\section*{Acknowledgment}
The authors used Anthropic's Claude, a large language model, to draft and revise
prose throughout the manuscript, to write the analysis code in
\texttt{analysis\_v2/} that computes the descriptors, the variance
decomposition, the precision and error-budget statistics and the figures, and to
help locate and verify the references. The study design, the acquisition of all
scan data, the interpretation of results and the conclusions are the authors'
own; the authors have verified every numeric claim against the computed result
tables and take full responsibility for the content.

\bibliographystyle{IEEEtran}
\bibliography{references}

\end{document}